\documentclass[conference]{IEEEtran}
\IEEEoverridecommandlockouts
\usepackage{cite}
\usepackage{amsmath,amssymb,amsfonts}
\usepackage{algorithmic}
\usepackage{graphicx}
\usepackage{textcomp}
\usepackage{xcolor}
\usepackage{hyperref}
\usepackage{multirow}
\usepackage{float}

\def\BibTeX{{\rm B\kern-.05em{\sc i\kern-.025em b}\kern-.08em
    T\kern-.1667em\lower.7ex\hbox{E}\kern-.125emX}}
\begin{document}
\title{Interaction-Aware Trajectory Prediction of Connected Vehicles using CNN-LSTM Networks}

\author{
\IEEEauthorblockN{Xiaoyu Mo}
\IEEEauthorblockA{\textit{School of Mechanical and}\\ \textit{Aerospace Engineering,} \\
\textit{Nanyang Technological University,}\\
\textit{Singapore 639798.} \\
xiaoyu006@e.ntu.edu.sg}
\and
\IEEEauthorblockN{Yang Xing}
\IEEEauthorblockA{\textit{School of Mechanical and}\\ \textit{Aerospace Engineering,} \\
\textit{Nanyang Technological University,}\\
\textit{Singapore 639798.} \\
xing.yang@ntu.edu.sg}
\and
\IEEEauthorblockN{Chen Lv*\thanks{Corresponding author: Chen Lv (e-mail: lyuchen@ntu.edu.sg).}}
\IEEEauthorblockA{\textit{School of Mechanical and}\\ \textit{Aerospace Engineering,} \\
\textit{Nanyang Technological University,}\\
\textit{Singapore 639798.} \\
lyuchen@ntu.edu.sg}
}
\maketitle
\begin{abstract}
Predicting the future trajectory of a surrounding vehicle in congested traffic is one of the basic abilities of an autonomous vehicle. In congestion, a vehicle's future movement is the result of its interaction with surrounding vehicles. A vehicle in congestion may have many neighbors in a relatively short distance, while only a small part of neighbors affect its future trajectory mostly. In this work, An interaction-aware method which predicts the future trajectory of an ego vehicle considering its interaction with eight surrounding vehicles is proposed. The dynamics of vehicles are encoded by LSTMs with shared weights, and the interaction is extracted with a simple CNN. The proposed model is trained and tested on trajectories extracted from the publicly accessible NGSIM US-101 dataset. Quantitative experimental results show that the proposed model outperforms previous models in terms of root-mean-square error (RMSE). Results visualization shows that the model is able to predict future trajectory induced by lane change before the vehicle operate obvious lateral movement to initiate lane changing.
\end{abstract}

\section{Introduction}
Human drivers always have a rough estimation of their surrounding vehicles' future movements, especially in congested traffic. They keep adjusting their next movement according to personal driving targets and the environment. It is important that an autonomous vehicle has the ability to predict future trajectories of its surrounding vehicles when sharing road with human drivers. This prediction is expected to be more precise than human drivers' estimation with rich information provided by connected vehicles, since human drivers can only roughly perceive the position of vehicles which are in sight. 
Connected vehicles based on reliable vehicle-to-vehicle (V2V) and vehicle-to-infrastructure (V2I) wireless communication are designed to improve the efficiency, response, and reliability of human driving and autonomous vehicles, while enhancing traffic safety and mobility. Connected vehicles technology enables real-time information sharing among surrounding vehicles and traffic management center~\cite{biswas2006vehicle, talebpour2016influence}. 

Even with enhanced information availability provided by connected vehicles, predicting a vehicle's future trajectory is challenging because it is affected by many factors, for example, the driver's orientation~\cite{schwarting2019social, xing2019personalized, xing2020energy , xing2020personalized}, different driving scenarios, and the interaction among vehicles~\cite{deo2018convolutional}. 

According to the survey~\cite{lefevre2014survey}, methods for trajectory prediction can be separated into three categories according to their degree of abstraction. Physics-based methods, which have the lowest level of abstraction, predict short term future trajectory of a vehicle based on its kinematic and dynamic properties~\cite{ammoun2009real}; Maneuver-based methods~\cite{hermes2009long, aoude2010threat, laugier2011probabilistic, althoff2009model}, which take into consideration the intention (maneuver) of a driver, are able to predict long term trajectories comparing to physics-based methods; Interaction-aware approaches~\cite{deo2018convolutional, ju2019interaction, zhao2019multi} consider the fact that the future trajectory of a vehicle is influenced by its surroundings and try to model this interaction for trajectory prediction. This nature of interaction-aware approach enables it to predict long term trajectories more precisely than other methods. 
\begin{figure*}
    \centering
    \includegraphics[trim={0cm 0cm 0cm 0cm}, clip, width=1.0\textwidth]{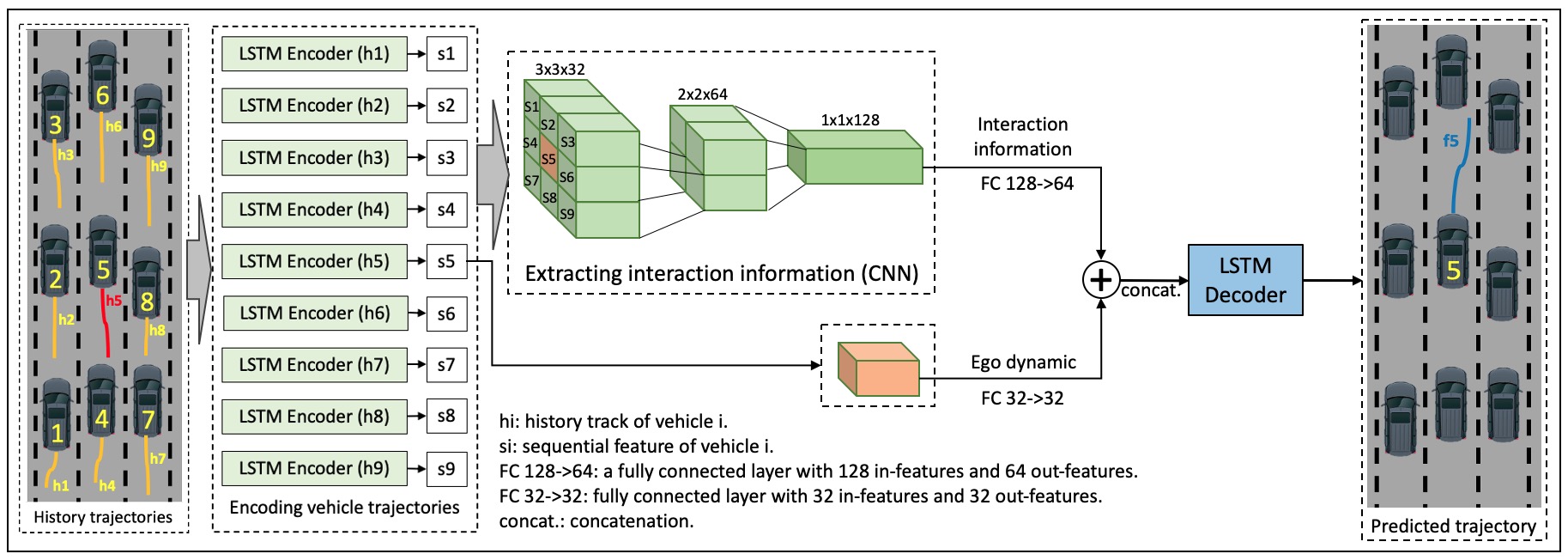}
    \vspace{-0.7cm}
    \caption{\textbf{Proposed model.} LSTMs with shared weights are used to encode the time serial information of each vehicle individually. A CNN based interaction extractor is then applied  to the LSTM encoding, which are implanted into a $3\times3$ grid according to direction. Finally a LSTM decoder takes as input the concatenated interaction and ego dynamic and outputs predicted trajectory of the ego.}
    \label{fig: model}
\end{figure*}

Among interaction-aware methods for trajectory prediction, the data-driven methods are prevalent because of the availability of traffic datasets~\cite{ushighway101, usi80freeway, wang2019apolloscape} and promising success of neural networks~\cite{krizhevsky2012imagenet}. 
Many of these methods are inspired by Social LSTM\cite{alahi2016social}, which uses long short-term memory networks (LSTMs)~\cite{hochreiter1997long} to encode dynamics of agents and models interaction by sharing  information among agents within a pre-defined distance. Social LSTM does not exploit the spatial structure of nearby agents and ignores vehicles in a longer distance. Convolutional social pooling~\cite{deo2018convolutional} considers the spatial structure by align LSTM encoded dynamics of neighboring vehicles into an ego centered $13\times3$ grid according to their local positions and applies a convolutional neural network (CNN)~\cite{krizhevsky2012imagenet} to extract interaction among neighboring vehicles excluding the ego. The final interaction is a combination of neighbors' interaction and dynamic of the ego. This model may redundantly consider vehicles which are in the grid but invisible to the driver in a congested traffic. Multi-agent tensor fusion (MATF)~\cite{zhao2019multi} takes as input a pixel level scene context and past trajectories of interacting agents and predicts trajectories of all agents in the scene. The spacial structure is retained by spatially aligning individual dynamics to pixel level context. SCALE-Net~\cite{jeon2020scale} uses edge-enhanced graph convolutional neural network (EGCN)~\cite{gong2019exploiting} and LSTMs to model inter-vehicle interaction and predicts trajectories of multiple vehicles. Interaction-aware Kalman Neural Networks (IaKNN)~\cite{ju2019interaction} uses a encoder-decoder structure to extract interaction-aware acceleration from rich environment observation. The environment observation includes sequences of accelerations, widths, lengths, relative distances and computed repulsive interaction forces of agents in the system.

Inspired by convolutional social pooling~\cite{deo2018convolutional} for vehicle trajectory prediction, a method with the same structure is proposed. The proposed model handles the interaction differently. As shown in Fig.~\ref{fig: model}, assuming that eight neighbors all exist and are perceptible to the ego in congestion, the proposed model uses two tubes to extract past dynamic of the ego and its interaction with eight neighboring vehicles separately. LSTM encoded dynamics of all 9 vehicles, including the ego, are aligned into an ego centered $3\times3$ grid according to their directions to the ego rather than their relative positions.  The $3\times3$ grid preserves surrounding vehicles' spacial structure because of coordinate system sharing described in sub-Sec.~\ref{subsec: problem}. A 2-layer CNN is designed to extract interaction feature hierarchically. Finally encoded ego dynamic and interaction are concatenated and feed to a LSTM decoder to predict future trajectory of the ego.

The proposed model is trained and tested on a dataset extracted from the publicly available dataset NGSIM US-101~\cite{ushighway101}. Evaluation results show that it is capable of predicting trajectories of vehicles and outperforms state-of-the-art methods. The major contributions of this work is summarized as below:
\begin{itemize}
    \item An interaction-aware trajectory prediction model is proposed;
    \item A method to choose roughly balanced lane-keeping and lane-changing scenarios from raw NGSIM dataset is implemented.
    \item Ablative studies are conducted on the extracted dataset to show the effectiveness of the proposed model.
\end{itemize}
The reminder of this paper is organized as below. Sec.~\ref{sec: model} formulates the problem and introduces the model in detail. Sec.~\ref{sec: data} describes how the data is processed. In Sec.~\ref{sec: results}, the proposed model is evaluated and compared to other models. Finally, this work is concluded in Sec.~\ref{sec: conclusion}.

\section{Methodology}
\label{sec: model}
\subsection{\textbf{Problem formulation}} 
\label{subsec: problem}
In this work, the task is to predict the trajectory of an ego vehicle based on history trajectories of its surrounding vehicles and its own. As shown in Fig.~\ref{fig: model}, eight surrounding vehicles and one ego (No.5) vehicle are considered. For each vehicle, its history is represented by a sequence of $16$ xy-coordinates of the past $3$ seconds. Past tracks of the ego and its surrounding vehicles are shown in triangular in red and gray, respectively. The ground truth of the future trajectory, which we want to predict as precisely as possible, is show in green. It is a $5$-second trajectory represented by $10$ points.

Surrounding vehicles considered are the ego's preceding (No.6) and following (No.4) vehicles, its nearest neighbors in adjacent lanes (No.2 and No.8), in terms of longitudinal distance, and their preceding (No.3 and No.9) and following (No.1 and No.7) vehicles. All $9$ history trajectories are aligned into a $3\times3$ grid according to directions. As in Fig.~\ref{fig: model}, history track of vehicle $i$ is allocated to the cell with number $i$ in the grid.

The input to the model is history trajectories:
\begin{equation}
    \mathcal{H}_t = \lbrace h^1_t, \cdots, h^9_t \rbrace,
    \label{eq: Hist}
\end{equation}
where
\begin{equation}
    \begin{aligned}
        h^i_t = & \lbrace (x^i_{t-n_h\cdot\Delta t_h}, y^i_{t-n_h\cdot\Delta t_h}), \\
                & (x^i_{t-(n_h-1)\cdot\Delta t_h}, y^i_{t-(n_h-1)\cdot\Delta t_h}), \\
                & \cdots, (x^i_t, y^i_t) \rbrace,
    \end{aligned}
\end{equation} is the track history of vehicle $i$ at time $t$. Each history $h^i_t$ is represented by $16=n_h+1$ xy-coordinates with time interval $\Delta t_h=0.2sec$. 

The output is future trajectory of the ego vehicle:
\begin{equation}
    \begin{aligned}
        f^5_t = & \lbrace (x^5_{t+1\cdot\Delta t_f}, y^5_{t+1\cdot\Delta t_f}), \\
                & (x^5_{t+2\cdot\Delta t_f}, y^5_{t+2\cdot\Delta t_f}), \\
                & \cdots, (x^5_{t+n_f\cdot\Delta t_f}, y^5_{t+n_f\cdot\Delta t_f}) \rbrace.
    \end{aligned}
\end{equation}
Ego's future trajectory is represented by $n_f=10$ xy-coordinates with time interval $\Delta t_f=0.5sec$. 

It is worth to note that all trajectories share the same coordinate system whose origin is fixed at position of the ego at time $t$ $(x^5_t, y^5_t)$. With this setting, trajectories of surrounding vehicles imply their relative positions with respect to the ego vehicle.

\subsection{\textbf{Model structure}}
\label{subsec: interaction}
The proposed model consists of two channels, one for the ego and the other one for the interaction between the ego and its surrounding vehicles. Individual and interaction information extracted from these two channels are then concatenated and taken as input by a LSTM decoder to predict future trajectory of the ego vehicle as shown in Fig~\ref{fig: model}.

\subsubsection{\textbf{LSTM encoder}}
A LSTM encoder is used for each vehicle to capture its individual bypast sequential feature. All vehicles share the same LSTM encoder. Eq.~\ref{eq: egolstm} shows the LSTM encoder applied to history of the ego $h^5_t$.
\begin{equation}
    E_t = \rm{FC_e}(\rm{LSTM_{enc}}(\rm{Emb}(h^5_t))),
    \label{eq: egolstm}
\end{equation}
where $\rm{Emb()}$ is a shared function embedding xy-coordinates into a higher space, $\rm{LSTM_{enc}()}$ is the shared LSTM encoder used in the proposed model, $\rm{FC_e()}$ is the a fully connected layer for the ego. In this channel,  
The ego's history track is embedded before sent to LSTM. The LSTM encoded feature is finally processed by a fully connected layer as the final representation of the ego's dynamic $E_t$. 

\subsubsection{\textbf{Interaction extractor}}
Individual sequential features extracted with LSTM should be jointly analyzed in order to capture the interdependence among vehicles. Social pooling~\cite{alahi2016social} addresses this issue by sharing information between spatially nearby LSTMs through a social pooling layer at each time step. In this setting, all agents within a certain distance are considered equally without exploiting the spacial structure. Convolutional social pooling~\cite{deo2018convolutional} defines an ego centered $13\times 3$ grid, which is populated with individual dynamics of surrounding vehicles according to their relative locations with respect to the ego. Then a 2-layer convolutional network is used to extract interaction among surrounding vehicles, considering the spacial information. If the grid is populated densely, it includes many vehicles, which cannot be perceived with on-board sensors by the ego, having no direct influence on the ego. In this work eight surrounding vehicles are implanted into an ego centered $3\times3$ grid according to their directions, rather than positions, with respect to ego. A 2-layer CNN is then applied to the grid to extract interaction among those vehicles, without introducing many vehicles having negligible impact on the ego in congestion. It seems that the proposed $3\times3$ grid discards the detailed relative position of surrounding vehicles. However, as we stated at the end of sub-Sec.~\ref{subsec: problem}, this spatial structure is inherently encoded by LSTMs because of coordinate system sharing.

In the first convolutional layer, $2\times2$ kernels are used to extract the interaction at four corners, upper left, upper right, lower left, and lower right. Each corner has four vehicles including the ego. The second convolutional layer also uses  $2\times2$ kernels to obtain the interaction among all vehicles by combining the interaction from four corners. The combined interaction is finally processed by a fully connected layer to be the final representation of inter-vehicle interaction. 

The proposed interaction extractor can be described by Eq.~\ref{eq: implant} and Eq.~\ref{eq: cnn}.
In Eq.~\ref{eq: implant}, history tracks of all vehicles $\mathcal{H}_t$ are encoded and then aligned ($\rm{Implant()}$) to a $3\times3$ grid as shown in Fig.~\ref{fig: model}.
Then this grid ($N^{lstm}_t$) is taken as input by the proposed 2-layer CNN ($\rm{CNN_{inter}()}$). Finally a fully connected layer $\rm{FC_N()}$ is used to summarize the final representation of interaction $N^{cnn}_t$.
\begin{equation}
    N^{lstm}_t = \rm{Implant}(\rm{LSTM_{enc}}(Emb(\mathcal{H}_t)))
    \label{eq: implant}
\end{equation}
\begin{equation}
    N^{cnn}_t = \rm{FC_N}(\rm{CNN_{inter}}(N^{lstm}_t)),
    \label{eq: cnn}
\end{equation}

\subsubsection{\textbf{LSTM decoder}}
Finally, a LSTM decoder ($\rm{LSTM_{dec}()}$) is used to generate predicted future trajectory $f^5_t$ of the ego vehicle. It takes as input the combination of interaction $N^{cnn}_t$ and ego's individual dynamic $E^t$ extracted by previous modules.
\begin{equation}
    f^5_t = \rm{LSTM_{dec}}([N^{cnn}_t, E^t]).
\end{equation}

\section{Data processing}
\label{sec: data}

\subsection{\textbf{Dataset}} 
The data used for training and testing is extracted from raw trajectories in NGSIM US-101~\cite{ushighway101}, which consists of vehicle trajectories on a segment, approximately 2,100 feet in length, of U.S. Highway 101. The trajectories are collected at 10 Hz between 7:50 a.m. and 8:35 a.m. on June 15, 2005. 
The study area includes five main lanes (lane 1 to 5), one auxiliary lane (lane 6), one on-ramp lane (lane 7), and one off-ramp lane (lane 8).
298 vehicles, which changed its lane for only once throughout the study area,  are selected as ego vehicles in this work. Further, trajectories before and after lane-changing are selected as data pieces, so that the resulted dataset includes roughly balanced lane-keeping and lane-changing scenarios. 

\subsection{\textbf{Data selection}}
The data used in this work is selected using a 2-step selection.
\subsubsection{\textbf{Lane change vehicles selection}} Vehicles satisfying following conditions are selected as ego vehicles:
\quad\begin{itemize}
    \item It has only been driving in lanes 1,2,3, and 4.
    \item Its lane ID has changed only once throughout the study area.
    \item The length of its trajectory is longer than 1,000 feet.
    \item Its longitudinal position, when changing lane, is within the range from 300 feet to 1,900 feet.
    \item The maximum lateral divergence of its trajectory from 6 seconds before lane change to 6 seconds after lane change is greater than 10 feet.
\end{itemize}
\subsubsection{\textbf{Data pieces selection}}
For a vehicle selected in the first step, 260 frames are considered as candidates of current frame (time $t$ in Eq.~\ref{eq: Hist}), which is the boundary of history and future trajectories. These frames are selected from 13 seconds before lane-change to 13 seconds after lane change. A data piece is selected if it meets conditions:
\begin{itemize}
    \item At time $t$, the ego has all 8 surrounding vehicles.
    \item The ego has a complete future trajectory with duration equals to 5 seconds.
    \item Each vehicle has a complete history with duration equals to 3 seconds.
\end{itemize}

With above two steps, 48150 data pieces are selected in total. 
The dataset is randomly split into two non-overlapping segments, 33705 ($70\%$) for training and 14445 ($30\%$) for testing.   

\section{Results and discussions}
\label{sec: results}
\subsection{\textbf{Metric}}
Root-mean-square error (RMSE) in meters of the predicted trajectories with regarding to the ground truth future trajectories is used to evaluate prediction accuracy of different models. RMSE is calculated for each predictive time step $t_p$ within a horizon of 5 seconds. The same metric was considered in previous works~\cite{deo2018convolutional, zhao2019multi, jeon2020scale}. 
\begin{equation}
    RMSE(t_p) = \sqrt{\frac{1}{n}\sum^{n}_{i=1}( (\hat{x}^i_{t_p} - x^i_{t_p} )^2 + ((\hat{y}^i_{t_p} - y^i_{t_p} )^2 )},
    \label{eq: metric}
\end{equation}
where $n=14445$ is the size of test set, $(\hat{x}^i_{t_p}, \hat{y}^i_{t_p})$ is the predicted position of the ego in data $i$ at time ${t_p}$, and $(x^i_{t_p}, y^i_{t_p})$ is the corresponding ground truth.

\subsection{\textbf{Implementation details}}
The model is implemented using PyTorch~\cite{NEURIPS2019_9015}.
Spacial coordinates are embedded into a 16-dimensional space before applying $\rm{LSTM_{enc}()}$ to encode individual past dynamics. The dimension of hidden states for $\rm{LSTM_{enc}()}$ and $\rm{LSTM_{dec}()}$ are 32 and 64, respectively. The 2-layer CNN uses fixed kernel size $2\times2$ but variant number of channels. The first layer has 32 in-channels and 64 out-channels, while the second layer has 64 in-channels and 128 out-channels. No padding is used in this model. 
$\rm{FC_e()}$ has 32 in-features and 32 out-features. $\rm{FC_N}$ has 128 in-features and 64 out-features. All layers have the same leaky-ReLU activation function with negative slope equals to $0.1$. An optimizer called ADAM~\cite{kingma2014adam} with learning rate of $0.001$ is used to train the model to minimize the weighted mean squared error (MSE) loss function defined in Eq.~\ref{eq: loss} on the extracted dataset. 
\begin{equation}
    \mathcal{L} = 20\times(\hat{X}-X)^2 + 0.5\times(\hat{Y}-Y)^2,
    \label{eq: loss}
\end{equation}
where $\hat{X}$ and $X$ are lateral positions of the predicted trajectory and the ground truth, $\hat{Y}$ and $Y$ are the corresponding longitudinal positions. More weight is given to lateral error considering the fact that lateral variance of a surrounding vehicle is much more important than its longitudinal position. And trajectories of vehicles driving on freeways almost always have much larger longitudinal movement than lateral movement.

\subsection{\textbf{Results}}
\label{subsec: results}
To demonstrate the advance of the proposed CNN-LSTM model, 4 models are implemented:
\begin{itemize}
    \item Vanilla LSTM (V-LSTM): this model uses a single LSTM to encode the individual history trajectory of the ego without considering the interaction among ego and its surrounding vehicles. 
    \item FC-LSTM: This is a variant of CNN-LSTM as described in sub-Sec~\ref{subsec: interaction}. The difference is that this model uses a fully connected layer to extract the interaction rather than a convolutional neural network.
    \item Interaction-only: This model discards the individual dynamic of ego in Eq.~\ref{eq: egolstm} and predicts ego's trajectory based only the interaction in Eq.~\ref{eq: cnn}.
    \item CNN-LSTM: This is the model proposed in this work. 
\end{itemize}

Above models are all trained and tested on the same dataset extracted from raw NGSIM US-101 trajectories with batch size equals to 8 for 20 epochs.
In addition, results of previous works~\cite{deo2018convolutional, zhao2019multi, jeon2020scale} are shown. It is worth to note that, although same metric is used through the implemented models and previous ones, the training and testing dataset are different from one work to another. 

\begin{table}[h!]
\caption{\textbf{Trajectory prediction results of diferrent models}}
\centering
\begin{tabular}{|c|c c c c c|} 
\hline
\multirow{2}{6em}{ \textbf{Methods }} &\multicolumn{5}{c|}{\textbf{Prediction horizon (Metric: RMSE in meters)}}\\
\cline{2-6} 
& \textbf{\textit{1 sec}}& \textbf{\textit{2 sec}}& \textbf{\textit{3 sec}} & \textbf{\textit{4 sec}} & \textbf{\textit{5 sec}} \\
\hline
CS-LSTM~\cite{deo2018convolutional} & 0.61 & 1.27 & 2.09 & 3.10 & 4.37 \\
\hline
SCALE-Net~\cite{jeon2020scale} & \textbf{0.459} & 1.156 & 1.973 & 2.911 & - \\
\hline
MATF GAN~\cite{zhao2019multi} & 0.66 & 1.34 & 2.08 & 2.97 & 4.13 \\
\hline
V-LSTM & 0.7393 & 1.7887 & 3.1321 & 4.8683 & 6.9017 \\
\hline
FC-LSTM & 0.657 & 1.0567 & 1.4399 & 1.9374 & 2.6296 \\
\hline
Interaction-only & 0.726 & 1.0193 & 1.3183 & 1.7247 & 2.4101 \\
\hline
CNN-LSTM & 0.6214 & \textbf{0.976} & \textbf{1.2751} & \textbf{1.6237} & \textbf{2.272} \\
\hline
\end{tabular}
\vspace{0.2cm}
\label{tab: othersresults}
\end{table}

Table~\ref{tab: othersresults} shows results of different models. Proposed interaction-aware models outperform previous models in long term trajectory prediction (2-5 seconds) in terms of RMSE in meters. This shows that dynamics of selected 8 surrounding vehicles contain enough information to model the interaction. 

The vanilla LSTM model has the poorest performance 
comparing to all listed interaction-aware trajectory prediction models. This indicates that modeling the interaction among vehicles is useful for trajectory prediction, even though the interaction can be modeled in different ways. This result is consistent with previous works~\cite{lee2017desire, deo2018convolutional, zhao2019multi, jeon2020scale}.

We note that, the proposed CNN-LSTM, which uses 2 convolutional layers to extract the interaction hierarchically, outperforms its variant FC-LSTM using fulle connected layers. This indicates that interaction is better modeled from local to global.

\begin{figure*}
    \centering
    \includegraphics[trim={4cm 0cm 0cm 0cm}, clip, width=1.0\textwidth]{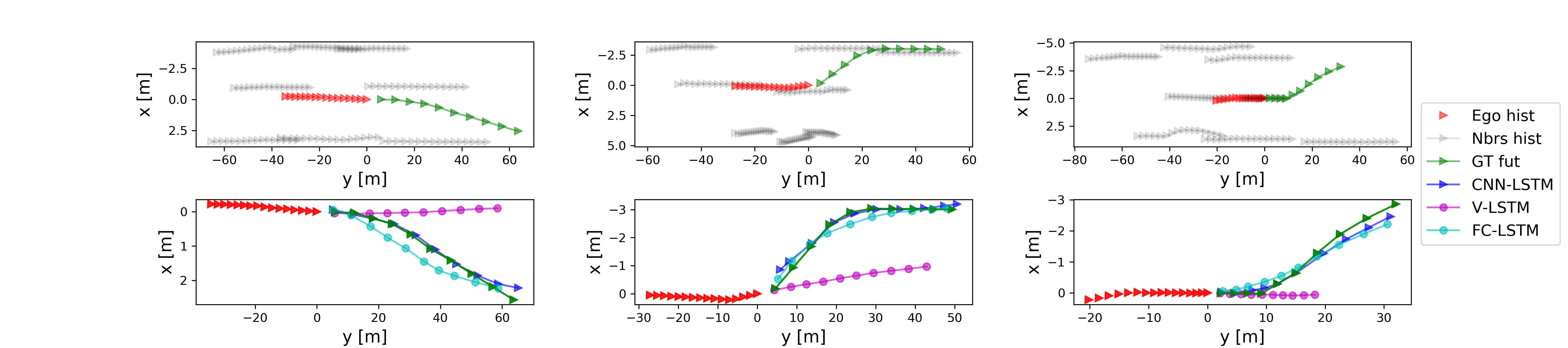}
    \vspace{-0.7cm}
    \caption{\textbf{Before lane change.} Trajectory prediction in driving scenarios before lane change, where x-axis represents lateral positions of vehicles in meters and y-axis represents their lateral positions. Trajectories are distinguished with colors as in the legend. Ego hist: history track of the ego; Nbrs hist: history tracks of surrounding vehicles; GT fut: Ground truth future trajectory of the ego; CNN-LSTM: Predicted trajectory of the proposed model; V-LSTM: Predicted trajectory of implemented V-LSTM; FC-LSTM: Predicted trajectory of implemented FC-LSTM.}
    \label{fig: beforelc}
\end{figure*}
\begin{figure*}
    \centering
    \includegraphics[trim={4cm 0cm 0cm 0cm}, clip, width=1.0\textwidth]{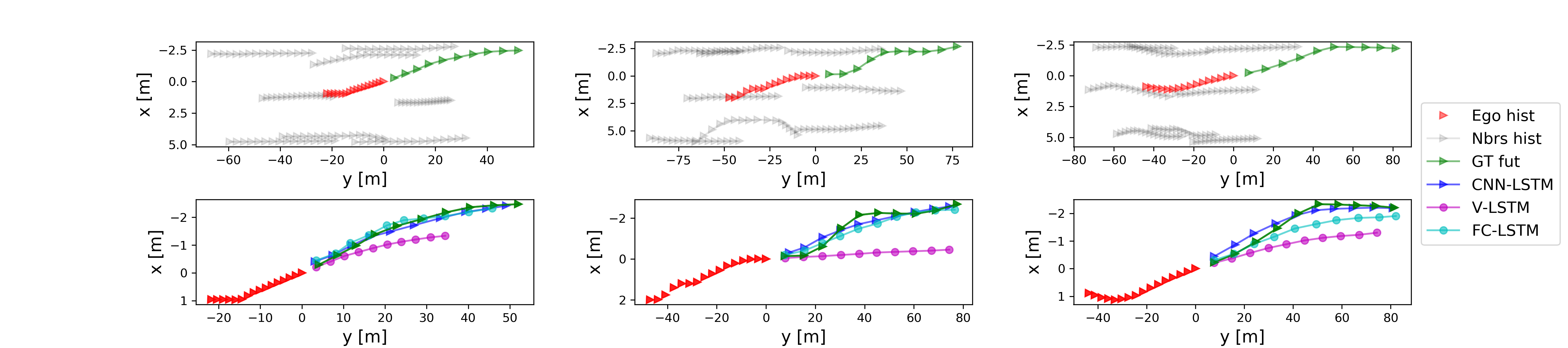}
    \vspace{-0.7cm}
    \caption{\textbf{During lane change.} Trajectory prediction in driving scenarios during lane change. Same legend is used here as in Fig.~\ref{fig: beforelc}.}
    \label{fig: duringlc}
\end{figure*}
\begin{figure*}
    \centering
    \includegraphics[trim={4cm 0cm 0cm 0cm}, clip, width=1.0\textwidth]{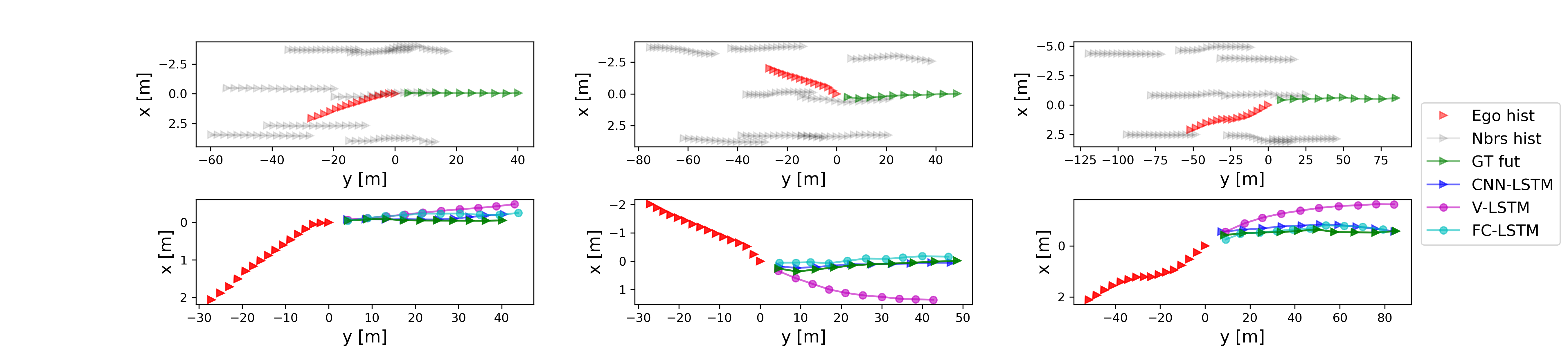}
    \vspace{-0.7cm}
    \caption{\textbf{After lane change.} Trajectory prediction in driving scenarios after lane change. Same legend is used here as in Fig.~\ref{fig: beforelc}.}
    \label{fig: afterlc}
\end{figure*}

The proposed CNN-LSTM also outperforms its variant, Interaction-only, where only the interaction $N^{cnn}_t$ in Eq.~\ref{eq: cnn} is used by the LSTM decoder. This shows that it is necessary to emphasize the ego's dynamic encoding with a separate tube, rather than just including it into interaction. 

\subsection{\textbf{Results visualization}}
Quantitative results in sub-Sec.~\ref{subsec: results} shows that the proposed model outperforms previous works and its variants in terms of RMSE in meters. In this section, predition results are visualized to study the performance of implemented models in different driving scenarios as shown in Fig.~\ref{fig: beforelc}, Fig.~\ref{fig: duringlc}, and Fig.~\ref{fig: afterlc}. For each figure, the first row shows three different driving scenarios, where all trajectory is shown in the same coordinate system. Since all history trajectories share the same time interval, their velocity can be roughly inferred from the dense of their trajectories. The second row shows prediction results of different models. 

Fig.~\ref{fig: beforelc} shows driving scenarios before lane change. It is clear that the vanilla LSTM model can hardly notice the ego's intention to change lane while interaction-aware models all predict trajectories reflecting the lane change intention. For example, the middle column of Fig.~\ref{fig: beforelc} shows a scenario where changing lane to left is a reasonable option. In the current lane, the preceding vehicle is slowing down; In the right lane, speeds of neighboring vehicles are much more slower than the ego, so that there is no reason to change to the right lane; In the left lane, speeds of preceding vehicles are faster than the ego and the following vehicle is slower. 

Fig.~\ref{fig: duringlc} and Fig.~\ref{fig: afterlc} show driving scenarios during and after lane change. The vanilla LSTM model cannot figure out whether the lane change maneuver is completed. However, interaction-aware models are able to know whether it is driving in center of the target lane (lane change completed) or between two lanes (during lane change) from its lateral distance to surrounding vehicles. 

\section{Conclusion}
\label{sec: conclusion}
In this work, an interaction-aware vehicular trajectory prediction method based on integrated CNN and LSTM is proposed for connected vehicles. In this model, LSTM encoders with shared weights are used to extract time serial information of individual vehicles; Then a CNN is applied to extract interaction among neighboring vehicles, whose sequential feature are implanted into a $3\times3$ grid according to their directions to the ego. Finally, the extracted interaction and ego's individual dynamic are concatenated and sent to a LSTM decoder to predict future trajectory of the ego vehicle. Quantitative results show that the proposed model outperforms existing works in terms of RMSE. Ablative studies on the selected dataset demonstrates the rationality of the proposed model. 

One limitation of the proposed model is that it assumes eight surrounding vehicles all exist and have a 3 second history. Future works could break this limitation by making a model adaptable to variational numbers of surrounding vehicles. Another way to improve this model is to improve it for multi trajectory prediction.  

\section*{Acknowledgment}
This work was supported by the SUG-NAP Grant (No. M4082268.050) of Nanyang Technological University, Singapore.
\bibliographystyle{ieeetr}
\bibliography{reference.bib}
\end{document}